\documentclass{article}

\PassOptionsToPackage{round}{natbib}

    \usepackage[preprint]{neurips_2025}

\usepackage[utf8]{inputenc} 
\usepackage[T1]{fontenc}    
\usepackage{hyperref}       
\usepackage{url}            
\usepackage{booktabs}       
\usepackage{amsfonts}       
\usepackage{nicefrac}       
\usepackage{microtype}      
\usepackage{graphicx}
\usepackage{booktabs}
\usepackage[table]{xcolor}
\usepackage{pifont}
\usepackage{algorithm}
\usepackage{algpseudocode}
\usepackage{lipsum}  

\usepackage{amsfonts}       
\usepackage{amsmath}        
\newcommand{\method}[1]{\textit{StatsMerging}}
\newcommand{\methodpp}[1]{\textit{StatsMerging++}}
\newcommand{\learner}[1]{\textit{StatsMergeLearner}}

\definecolor{darkcyan}{rgb}{0.0, 0.55, 0.55}
\newcommand{\bo}[1]{\textcolor{darkcyan}{(Bo):{#1}}}

\definecolor{darkgreen}{RGB}{0,100,0}
\newcommand{\cmark}{\textcolor{darkgreen}{\checkmark}}
\definecolor{darkred}{RGB}{139,0,0}
\newcommand{\xmark}{\textcolor{darkred}{\ding{55}}}
\newcommand{\cmarkk}{\checkmark}

\definecolor{gray}{rgb}{0.3, 0.3, 0.3}
\definecolor{darkblue}{rgb}{0.0, 0.0, 0.55}
\definecolor{lightblue}{rgb}{0.68, 0.85, 0.9}

\hypersetup{
    colorlinks=true,       
    linkcolor=black,        
    citecolor=darkblue, 
    filecolor=black, 
    urlcolor=darkblue
}

\usepackage{comment}
\usepackage{enumitem}

\title{Statistical Structure-Aware Vision Model Merging}
\title{Model Weights Statistical Geometry-Aware Vision Model Merging}
\title{Statistical Geometry-Aware Merging of Vision Models}
\title{Vision Model Merging Guided by Statistical Geometry of Weights}
\title{Vision Model Merging through Statistical Geometry of Weights}
\title{Vision Model Merging through Adaptive MergeLearner}
\title{\method~: Unsupervised Stats-Guided Model Merging via MergeLearner (Not unsupervised)}
\title{\method~: Stats-Guided Model Merging using Multi-Teacher Distillation Pseudo Labels}
\title{\method~: Stats-Guided Model Merging via Task-Teacher Distillation}
\title{\method~: Statistics-Guided Model Merging via Task-Specific Teacher Distillation}

\author{%
  Ranjith Merugu\textsuperscript{*} \quad
  Bryan Bo Cao\textsuperscript{*} \quad
  Shubham Jain \\
  Department of Computer Science, Stony Brook University \\
  Stony Brook, NY 11790 \\
  \texttt{\{rmerugu,boccao,jain\}@cs.stonybrook.edu} \\
  \textsuperscript{*}Equal contribution
}

\begin{document}

\maketitle

\begin{abstract}
Model merging has emerged as a promising solution to accommodate multiple large models within constrained memory budgets. We present \method~, a novel lightweight learning-based model merging method guided by weight distribution statistics without requiring ground truth labels or test samples. \method~ offers three key advantages: (1) It uniquely leverages singular values from singular value decomposition (SVD) to capture task-specific weight distributions, serving as a proxy for task importance to guide task coefficient prediction; (2) It employs a lightweight learner \learner~ to model the weight distributions of task-specific pre-trained models, improving generalization and enhancing adaptation to unseen samples; (3) It introduces \textit{Task-Specific Teacher Distillation} for merging vision models with heterogeneous architectures, a merging learning paradigm that avoids costly ground-truth labels by task-specific teacher distillation. Notably, we present two types of knowledge distillation, (a) distilling knowledge from task-specific models to \learner~; and (b) distilling knowledge from models with heterogeneous architectures prior to merging. Extensive experiments across eight tasks demonstrate the effectiveness of \method~. Our results show that \method~ outperforms state-of-the-art techniques in terms of overall accuracy, generalization to unseen tasks, and robustness to image quality variations.
\end{abstract}

\section{Introduction}
\label{sec:introduction}


Computer vision has witnessed transformative progress fueled by deep learning, particularly through the development and adoption of large-scale pre-trained models. Architectures like Convolutional Neural Networks (CNNs) \citep{krizhevsky2012imagenet, he2016deep, simonyan2014very}, Vision Transformers (ViTs) \citep{dosovitskiy2020image, touvron2021training}, and hybrid approaches \citep{liu2022convnet} pre-trained on massive datasets 
have become the cornerstone of modern vision applications. Large-scale models leveraging multi-modal pre-training, such as CLIP \citep{radford2021learning}) or generative models like GANs \citep{goodfellow2014generative} and Diffusion Models \citep{ho2020denoising, rombach2022high} have further pushed the boundaries of visual understanding and synthesis, enabling the use of pre-trained backbones to a wide range of downstream vision applications. Fine-tuning these powerful base models has become the dominant practice in a wide range of computer vision tasks. This success, however, leads to a practical challenge: the proliferation of numerous specialized pre-trained weights and model checkpoints \citep{cao2025few}, most of which share the same foundational ViT or CNN backbones. Managing this growing collection incurs significant storage overhead, complicates deployment, and represents a missed opportunity to consolidate the related, yet specialized, knowledge contained within these models \citep{wortsman2022model}, particularly on compute-constrained platforms such as edge devices \citep{cao2024representation, singh2024ovida}. While Multi-Task Learning (MTL) \citep{vandenhende2021multi, zhang2021survey} aims to create versatile single models for vision tasks, it often demands complex joint training strategies, concurrent access to diverse datasets, and careful architecture design to balance performance across disparate tasks.

Model merging offers a compelling post-hoc alternative, seeking to combine independently trained models without expensive retraining. However, while techniques for model merging have gained traction, particularly in Natural Language Processing (NLP) \citep{yadav2023ties, ilharcoediting}, adapting these techniques in computer vision domain has far less explored. A straightforward approach of simple weight averaging \citep{wortsman2022model} often fails in vision tasks due to the complex, hierarchical visual feature representations, task-specific optimizations, and the presence of intricate noise patterns that lead to sharp, non-convex loss minima \citep{izmailov2018averaging}. Recent methods in this direction \citep{matena2022merging, jindataless, yang2023adamerging, padmanabhan2023gemel} neglect the importance of weight distribution.

\begin{figure}[h]
  \centering
    {\includegraphics[width=1\textwidth]{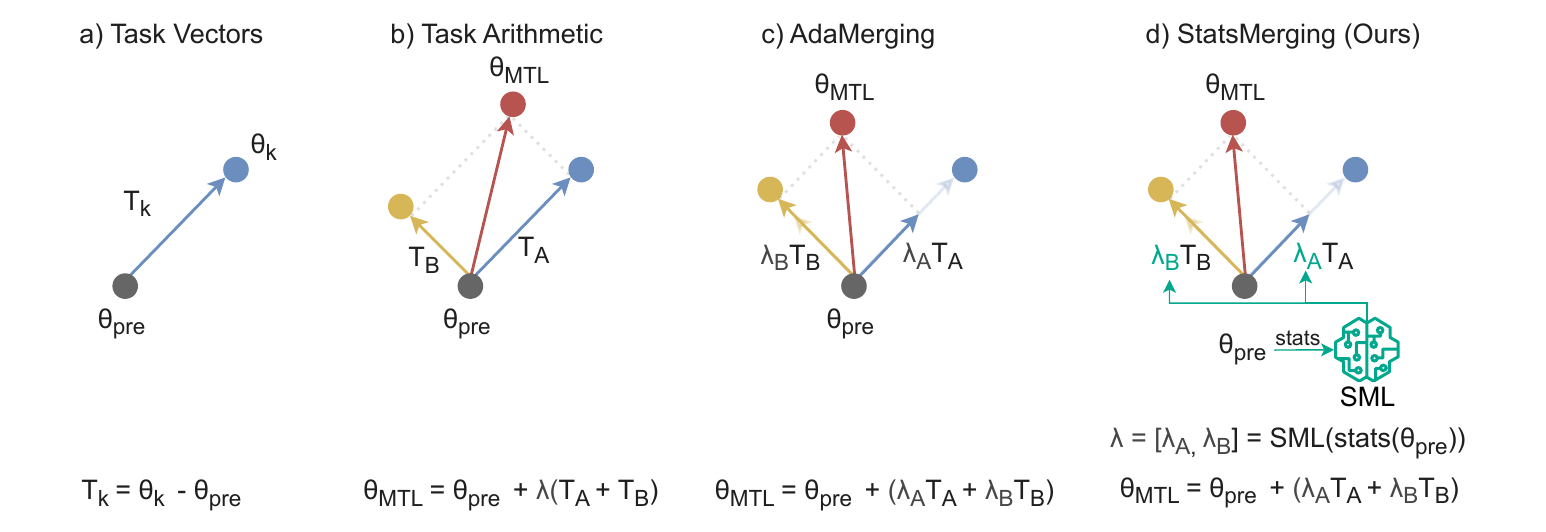}}
\vspace{-16pt}
  \caption{Compared to prior works, \method~ uniquely learns the merging coefficients using \learner~, taking advantage of statistical features of weigts pre-trained on prior tasks. Notably, while both AdaMerging and \method~ are presented in the task-wise level in c) and d) for simplicity of illustration, the same principle can be applied at the layer-wise level for fine-grained adaptation.
  }
  \label{fig:statsmergingvsothers}
\end{figure}

This paper introduces a novel model merging framework specifically designed to address the aforementioned challenges within computer vision. We propose \method~, a weight distribution statistics-guided merging approach that moves beyond simple parameter averaging or task-vector manipulation. \method~ leverages the statistical features models pre-trained on prior tasks for merged. In particular, we compute salient statistics extracted by leverage Singular Value Decomposition (SVD) to capture the dominant properties of the learned feature spaces. This statistical information, intrinsically capturing aspects of the pre-trained model distributions, guides the merging process by learning a compact Multilayer Perceptron (MLP), coined \learner~ that predicts adaptive merging coefficients ($\lambda$) shown in Fig. \ref{fig:statsmergingvsothers}. This allows the merging to be guided by the weight landscape, rather than treating coefficients as free parameters requiring external tuning data.

\vspace{-6pt}

We make four significant contributions summarized as follows:
\begin{itemize}
    \item We propose \method~\footnote{Our code is available at \href{https://github.com/statsmerging/statsmerging}{https://github.com/statsmerging/statsmerging.}}, a novel model merging framework guided by model weight statistics, leveraging singular values extracted via Singular Value Decomposition (SVD) to predict merging coefficients $\lambda$.

    \item We design the lightweight \learner~ to learn model merging coefficients $\lambda$ estimation based on weight statistical features, through a newly proposed Task-Specific Teacher Distillation paradigm without manually-annotated labels.

    \item We introduce the first heterogeneous architectural merging method, which distills knowledge from models with non-identical architectures into the unified target architecture.

    \item Extensive experiments demonstrate the effective of our proposed \methodpp~, an extended version of \method~, which achieves $94.5\%$ average accuracy on merging models from eight tasks, outperforming the state-of-the-art WEMoE ($89.4\%$) by a substantial margin of $5.1\%$.
\end{itemize}

\section{Related Work}
\subsection{Multi-Task Learning}
Multi-Task Learning (MTL) \citep{vandenhende2022multi} represents a paradigm for training a single model to perform multiple tasks concurrently. While MTL aims to create unified models capable of handling diverse objectives, it typically requires careful design of network architectures, computationally expensive training, access to large and diverse datasets, and intricate task balancing strategies \citep{zhang2021survey}. Model merging offers a compelling alternative by enabling the combination of independently trained models, without the need for extensive retraining or simultaneous access to multi-task datasets or privacy-restricted data \citep{jindataless}.

\subsection{Multi-Task Merging}
Early approaches to model merging often involved simple heuristics like Weight Averaging \citep{wortsman2022model}, Ties-Merging \citep{yadav2023ties}, and Arithmetic Merging \citep{ilharcoediting}. While straightforward to implement, these methods typically lack awareness of the weight distributions and learned representations within the models, leading to suboptimal performance in the merged model compared to individually fine-tuned models or unified models trained from scratch. \citep{wortsman2022model} demonstrate that naive weight averaging could significantly degrade performance, highlighting the challenges in consolidating knowledge from independently trained networks. Methods explored in natural language processing \citep{yadav2023merging, ilharcoediting} have also shown promise by learning interpolation weights.

More recent efforts in model merging have introduced various strategies to efficiently combine multiple models without retraining. Approaches such as ZipIt \citep{zhang2024zipit}, EMR-Merging \citep{huang2024emr}, and Training-Free Pretrained Model Merging methods \citep{sun2025cat, chen2024retraining} emphasize data-free, tuning-free methodologies, often leveraging weight-space heuristics or task-vector alignment. Techniques like Pareto Merging \citep{chen2025pareto}, MAP \citep{li2024map}, and $C^{2}M^{3}$ \citep{crisostomi2024c} formulate model merging as a multi-objective or constrained optimization problem to preserve task performance across domains. Other works such as Parameter Competition Balancing \citep{du2024parameter} and Sharpness-Aware Fine-Tuning \citep{lee2025mitigating} address parameter interference during merging. Meanwhile, methods like LayerMerge \citep{kim2024layermerge} and MERGE3 \citep{mencattini2025merge} aim to improve scalability and computational efficiency, making merging feasible on consumer-grade hardware. WEMoE \citep{tang2024merging} ensembles shared and task-specific MLPs with input-conditioned routing in a layer-wise, data-free manner. Representation Surgery \citep{yang2024representation} introduces a scheme to alleviate the problem of representation bias while Evolutionary Model Merge \citep{akiba2025evolutionary} employs evolutionary algorithms to optimize model merging recipes.

These methods, however, do not explicitly leverage the weight distribution of the models being merged, a key distinction from our proposed approach. The gap often lies in effectively unifying the diverse and task-specific feature representations learned by individual models into a single, high-performing entity without extensive learning.

\subsection{Merging Methods in Computer Vision}
The application of model merging techniques in computer vision is relatively less explored compared to natural language processing \citep{yadav2023merging, ilharcoediting}. Computer vision models, particularly deep convolutional neural networks (CNNs) \citep{krizhevsky2012imagenet, he2016deep, simonyan2014very} and Vision Transformers (ViTs) \citep{dosovitskiy2021an, touvron2021training}, learn complex, hierarchical feature representations that are highly sensitive to task-specific optimizations \citep{izmailov2018averaging}. Simple averaging techniques often fail due to the non-convex nature of the loss landscape and the divergence of learned feature spaces across different visual tasks. Recent advancements \citep{matena2022merging, yang2023adamerging} have shown potential, but often lack explicit mechanisms to account for the unique properties inherent in visual data and architectures, such as spatial relationships in CNNs \citep{cao2023data, cao2024lightweight} or attention mechanisms in ViTs \citep{ye2023merging, tang2025merging}. Our work addresses these limitations by introducing a novel merging framework that leverages internal model weight statistics to guide the merging process, making it more adaptable and effective across diverse computer vision tasks and architectures.

\vspace{-5pt}
\begin{table*}[h]
\footnotesize
  \begin{center}
    {
\begin{tabular}{c|cccccc}
\toprule
Method & No & Layer & TT & Heterogeneous \\
  & Manual Label & Level & Adaptability & Architecture \\
  \midrule
 Traditional MTL & \xmark & * & \xmark & \xmark \\
\midrule
 Task Arithmetic & \cmark & \xmark & \xmark & \xmark \\
 Ties-Merging & \cmark & \xmark &  \cmark & \xmark \\
 Fisher Merging & \cmark & \xmark & \xmark & \xmark \\
 RegMean & \cmark & \xmark & \xmark & \xmark \\
 EMR-MERGING & \cmark & \cmark & \xmark & \xmark \\
  AdaMerging & \cmark & \cmark & \cmark & \xmark \\
  Representation Surgery & \cmark & \cmark & \cmark & \xmark \\
  WEMoE & \cmark & \cmark & \cmark & \xmark \\
 \midrule
 \rowcolor{gray!15} \method~ (Ours) & \cmark & \cmark & \cmark & \cmark \\
\bottomrule
\end{tabular}
}
\end{center}
\vspace{-5pt}
\caption{Summary of system characteristics in recent works. *: Optional. TT: Test-Time.} 
\label{tab:pk}
\end{table*}

In summary, our method \method~ enjoys several advantages compared to prior works shown in Table \ref{tab:pk}: (1) no human annotated labels are required for weight distribution learning; (2) It operates at a fine granularity, specifically at the layer-wise level; (4) it allows for test-time adaptability; (5) it facilitates extension to heterogeneous architectures.

\section{Methodology}


\subsection{Preliminaries}
\noindent \textbf{Notations: } A deep neural network is parameterized by a set of weights $\theta = \{\theta_{1}, \theta_{2}, \ldots, \theta_{L}\}$ that learns the mapping from an input data $x_{i} \in \mathbb{R}^{d}$ to a predicted value $\hat{y_{i}} \in \mathbb{R}^{D}$: $f_{\theta}(x_{i}) \rightarrow \hat{y_{i}}$. Of these, $\theta^{l}$ represents the $l$-th $l \in \{1, 2, \ldots, L\}$ layer weights where $L$ is the number of layers of the model $f_{\theta}$, $d$ denotes an input data $x_{i}$'s dimension. For classification problems, $y_{i}$ is the class label and $D$ is the number of classes, while for regression problems, $D$ is the dimension of the output vector $y_{i}$.

The weights of a pre-trained model (e.g., ViT or ResNet) are denoted as $\theta_{pre} = \{\theta_{pre}^{1}, \theta_{pre}^{2}, \ldots, \theta_{pre}^{L}\}$. The weights fine-tuned on a specific training data $\{x_{i}, y_{i}\}^{N^{tr}_{k}}_{i=1}$ for task $k$ is recorded as $\theta_{k} = \{\theta_{k}^{1}, \theta_{k}^{2}, \ldots, \theta_{k}^{L}\}$ where $N^{tr}_{k}$ is the number of training samples.

\noindent \textbf{Problem Formulation: } The problem of \textit{model merging} is formulated as given $K$ tasks' training data, find a way to combine weights $\{\theta_{k}\}_{k=1}^{K}$ fine-tuned for $K$ tasks previously to obtain a new weight $\theta_{m}$ without undergoing the retraining process, while the new model $f_{\theta_{m}}$ is capable of performing well on $K$ tasks jointly.

It is assumed that all $K$ fine-tuned weights and the merged weight share the same neural network architecture. Therefore, the core question is how to \textit{linearly combine} $\{\theta_{k}\}_{k=1}^{K}$ to obtain $\theta_{m}$. In the task level, the model merging problem is finding a set of coefficients $\lambda_{k} \in \{\lambda_{1}, \lambda_{2}, \ldots, \lambda_{K}\}$ such that the merged model weights $\theta_{m} = \sum_{k=1}^{K}\lambda_{k}\theta_{k}$ for model $f_{\theta_{m}}$ perform well on all $K$ tasks. In the layer level, it becomes searching for a set of coefficients $\lambda_{k}^{l} \in \{\lambda_{1}^{1}, \lambda_{1}^{2}, \ldots, \lambda_{1}^{L}, \lambda_{2}^{1}, \lambda_{2}^{2}, \ldots, \lambda_{2}^{L}, \ldots, \lambda_{K}^{L}\}$ to obtain the merged model $\theta_{m} = \sum_{k=1}^{K}\sum_{l=1}^{L}\lambda_{k}^{l}\theta_{k}^{l}$ that maintain high performance on $K$ tasks.

\subsection{Weight Statistics-Guided Model Merging}

In this section, we describe the main intuition and techniques of our proposed method: \textit{\method~}. Our core idea is that given the distribution of pre-trained weights $\theta_{k}$, we can learn a function $g(\theta_{k}) \rightarrow \lambda_{m}$ to predict the merging coefficients $\lambda_{m}$. We argue that \textit{weight distribution} plays an important role in model merging. However, directly using the raw weights $\theta_{k}$ as input is impractical due to the high dimension of $\theta_{k}$. We posit that such information can be represented by weight statistics. These statistical features contain key information regarding the amount of weights $\theta_{k}$ for a task $k$ to be merged to the final model. We highlight the key differences with prior works in Fig. \ref{fig:statsmergingvsothers}.

\noindent \textbf{Weight Statistics: } For a pre-trained weight $\theta_{k}$ on task $k$, we compute the mean $\mu_{\theta_{k}}$ and variance $\sigma^2 = Var(\theta_{k})$ to represent its center and breadth, as well as its magnitude $m=||\theta_{k}||$. The underlying intuition is based on the observation that merging performance is largely influenced by high-magnitude parameters \citep{yadav2023ties}. In addition, we extract the singular values $\sigma^{\prime}_{i}$ from Singular Value Decomposition (SVD):
\begin{equation}
    W_{{k}} = U_{{k}}\Sigma_{{k}} V^\top_{{k}}
\end{equation}
where $W_{\theta_{k}}$ represents the matrix of the model parameter $\theta_{k}$.
By default, we use rank $3$ from $\Sigma_{{k}}$ to form weight statistics. Motivated by prior findings on the effectiveness of SVD in neural network pruning \citep{goetschalckx2018efficiently, abid2002new, kim2025singular}, we hypothesize that singular values encapsulate essential information regarding the weight distribution, which can guide the allocation of weights from $\theta_{k}$ during merging.

Combining all together, the weight statistics feature vector $S_{{k}}$ is formed as
\begin{equation}
    S_{{k}} = stats(\theta_{k}) = [\mu, \sigma^{2}, m, \sigma_{r}^{\prime}]
    \label{equ:task_S}
\end{equation}
where $stats()$ extracts the statistical features from the weight $\theta_{k}$, $\sigma_{r}$ represents the singular value vector given rank $r$: $\sigma_{r}^{\prime}=[\sigma_{1}^{\prime}, \sigma_{2}^{\prime}, \ldots, \sigma_{r}^{\prime}]$.

Notably, the Equation \ref{equ:task_S} above is task-wise while we also introduce layer-wise formulation for layer $l$:
\begin{equation}
    S_{k}^{l} = stats(\theta_{k}^{l}) = [\mu, \sigma^{2}, m, \sigma_{r}^{\prime}]^{l}
    \label{equ:task_S}
\end{equation}
where the layer-wise statistics features of pre-trained model from task $k$ layer ${l}$ is computed.

\noindent \textbf{StatsMergeLearner (SML): } We adopt a multilayer perceptron (MLPs) to learn to predict the merging coefficients $\lambda$ given weight statistics feature vector $S_{{k}}$ as input. In the task-wise mode, the \learner~ is denoted as $SML(S_{{k}})$:
\begin{equation}
    \lambda_{k} = SML(S_{{k}}) = g(stats(\theta_{k}))
\end{equation}
where $\lambda_{k}$ is a scalar representing the merging coefficient of Task $k$ model.
In the layer-wise mode, the \learner~ is denoted as $M(S_{{k}})$:
\begin{equation}
    \lambda_{k}^{l} = SML(S_{k}^{l}) = g(stats(\theta^{l}_{k}))
\end{equation}
    where $\lambda_{k}$ is a vector containing $L$ layers' coefficients and $\lambda_{k}^{l}$ refers to the coefficient of layer $l$ in the $k$ pre-trained model. By default, we use a two-layer MLP to implement the \learner~.

\noindent \textbf{Optimization Objective.}
To train \learner~, in the standard supervised training paradigm, we freeze the weights for each task $\theta_{k}$ and apply the cross-entropy loss function $L_{CE}$ on the aggregated dataset:
\begin{equation}
    \mathcal{L}_{\mathrm{CE}}^{SL} = - \sum^{C_{m}}_{c=1} y_{c} \log(\hat{y}_{c}))
\end{equation}
where $\hat{y}_{c}$ is the prediction from the merged model for class $c$, $C_{m}$ is the total number of classes in the aggregated dataset.

\subsection{Task-Specific Teacher Distillation}
\label{sec:ttd}

\begin{figure}[h]
\hspace*{-18pt}
  \centering
    {\includegraphics[width=1.1\textwidth]{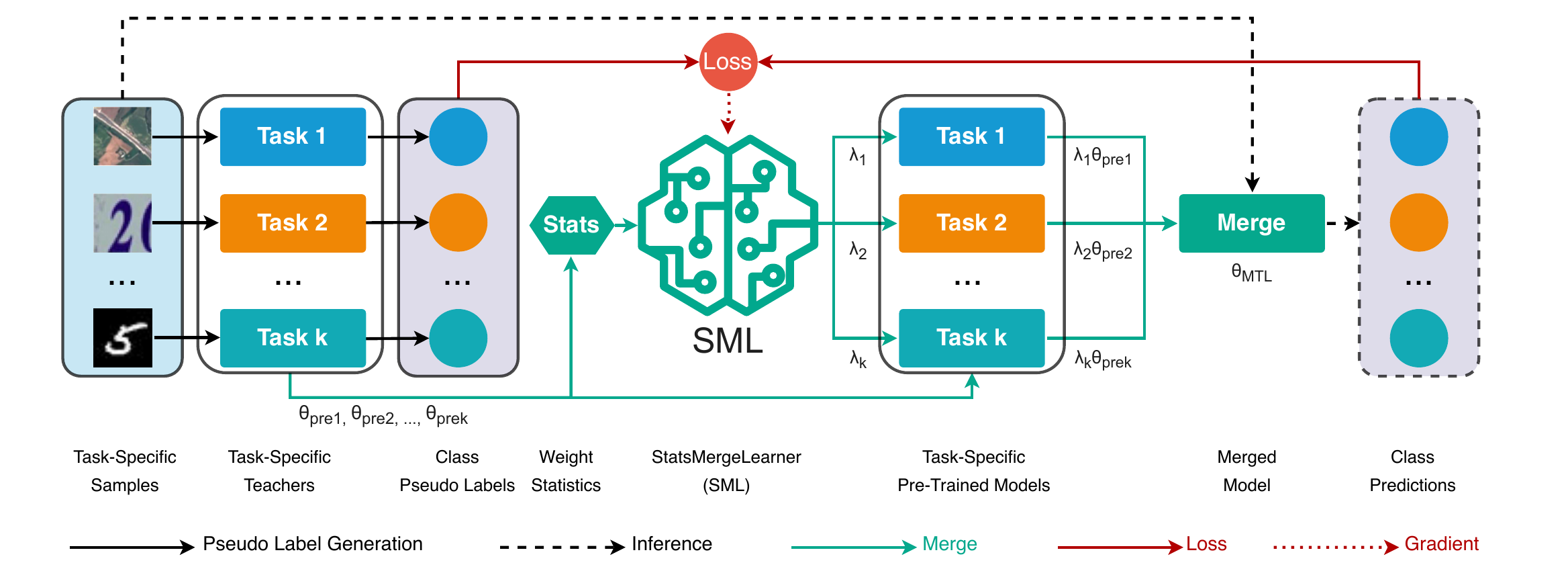}}
\vspace{-15pt}
  \caption{Knowledge Distillation Diagram. \learner~ (SML) learns the merging coefficients $\lambda$ by minimizing the loss between the merged model’s predictions and pseudo labels generated by task-specific teacher models. During inference, only the merged model in \method~ is used to predict class labels.
  }
  \label{fig:kd}
\end{figure}

We present a novel Task-Specific Teacher Distillation training paradigm to train the \learner~ (SML) for model merging as illustrated in Fig. \ref{fig:kd} and detailed in Algorithm 1. 


\noindent
\begin{minipage}[t]{0.53\textwidth}
\raggedright

Our key intuition is that each pre-trained model $\theta_{k}$ is already good at its own task dataset $\{x_{i}, y_{i}\}_{k} \ \in D_{k}$, therefore we regard it ($\theta_{k}$) as the Task-Specific Teacher $T_{k}$. Subsequently, the predictions $\hat{y}_{i, k}$ from the model trained on task $k$ serves as sufficiently reliable pseudo labels for the validation dataset sample $\{x_{i}, y_{i}\}_{k}$ from the same task. We aggregate such pairs $\{x_{i}, \hat{y}_{i,k}\}_{k}$ to construct the merged dataset to train $SML()$. It is important to note that all samples for $SML()$ learning are collected from the validation set only. We highlight the key benefit of this approach that enables dataset preparation without relying on human-annotated labels. The predicted class label in one-hot encoded format. Therefore, the cross-entropy loss is applied:
\begin{equation}
    \mathcal{L}_{\mathrm{CE}} =  - \sum^{C_{m}}_{c=1} \hat{y}_{c,k} \log(\hat{y}_{c})).
\end{equation}

\end{minipage}%
\hfill
\begin{minipage}[t]{0.43\textwidth}
\scriptsize
{Algorithm 1. }{Unified Statistics-Guided Model Merging via Task-Specific Teacher Model Distillation}
\begin{algorithmic}[1]
\State \textbf{Input:} Set of pre-trained models $\{M_1, M_2, \ldots, M_k\}$ with weights $\{\theta_1, \theta_2, \ldots, \theta_k\}$ for $K$ tasks.
\State \textbf{Output:} Merged model $M_{\text{merged}}$ with weights $\theta_{\text{merged}}$
\State // Prepare $K$ pre-trained models
\If{Same architecture $A$ for all $M_i$}
  \State Set $M_{\text{target}}$ to the shared architecture
\Else
  \State Select a target architecture $M_{\text{target}}$
  \For{$i = 1$ to $k$}
    \If{$A(M_i) \neq A(M_{\text{target}})$}
      \State Distill $M_i$ into $M_{\text{target}}$ to obtain updated $\theta_i$
    \EndIf
  \EndFor
\EndIf
\State // Merge $K$ models
\For{$k = 1$ to $K$}
  \State // mean $\mu$, std $\sigma^{2}$, norm $m$, singular value $\sigma_{r}^{\prime}$
  \State Extract statistics $S_k = [\mu, \sigma^2, m, \sigma_r']$ from $\theta_k$
  \State Predict coefficients $\lambda_k = \text{SML}(S_k)$
  \State Merge layer weights: $\theta^l_{\text{merged}} = \sum_{i=1}^k \lambda_k \theta_k$
\EndFor
\State \Return $M_{\text{merged}}$ with weights $\theta_{\text{merged}}$
\end{algorithmic}
\label{algo:distill}
\end{minipage}

The use of a simple cross-entropy loss function allows for extending to other vision tasks and architectures detailed in Sec. \ref{sec:mperformance}.

\section{Experiments}
\vspace{-10pt}
\subsection{Experimental Setup}

In this section, we present the experimental setup following AdaMerging \citep{yang2023adamerging} and evaluation results used to compare our method against recent baselines.

\noindent \textbf{Datasets and Models}:
Our experiments include eight image classification tasks with datasets SUN397 \citep{xiao2016sun}, Stanford Cars \citep{krause20133d}, RESISC45 \citep{cheng2017remote}, EuroSAT \citep{helber2019eurosat}, SVHN \citep{netzer2011reading}, GTSRB \citep{stallkamp2011german}, MNIST \citep{lecun1998mnist}, DTD \citep{cimpoi2014describing}, and CIFAR10 \citep{krizhevsky2009learning} \footnote{In the remainder of the paper, the abbreviations shown in brackets are used to denote each task dataset: SUN397 (SU), Cars (CA), RESISC45 (RE), EuroSAT (EU), SVHN (SV), GTSRB (GT), MNIST (MN) and DTD (DT).} We use ViT-B/32 CLIP \citep{radford2021learning} as the pre-trained backbone. Individual task-specific models are obtained by training on each dataset separately. For merging models with different architectures, we first distill them into a single backbone before applying our merging method.



\noindent \textbf{Baselines and Metrics}:
We compare against standard baselines including Individual Training, Traditional Multi-Task Learning (MTL) \citep{zhang2021survey}, Weight Averaging \citep{wortsman2022model}, Task Arithmetic \citep{ilharcoediting}, Fisher Merging \citep{matena2022merging}, RegMean \citep{jindataless}, Ties-Merging \citep{yadav2023ties}, EMR-MERGING \citep{huang2024emr}, AdaMerging \citep{yang2023adamerging}, Representation Surgery \citep{yang2024representation}, SurgeryV2 \citep{yang2024surgeryv2}, and WEMoE \citep{tang2024merging}. The primary evaluation metric is the average accuracy (Avg Acc) on the test sets of all tasks. The evaluation is conducted on eight different vision classification tasks.


\noindent \textbf{\learner~ Training Detail}:
Our MLP-based \learner~ learns to predict layer-wise or task-wise merging weights coefficients ($\lambda$) based on weight statistics from individual task models. The \learner~ is trained for 500 epochs using Adam, with a learning rate of $1e-3$ and a StepLR scheduler (factor 0.1 every 100 epochs), which translates to around only 3 hours to merge 4 ViTs, offering the practicality and advantage of applying our technique for practitioners without spending days or weeks for training \citep{zhang2021survey, padmanabhan2023gemel}. We train the \learner~ primarily using knowledge distillation from the aggregated dataset without human annotated labels described in Sec. \ref{sec:ttd}, optimized with either Cross-Entropy \citep{mao2023cross} or KL Divergence loss \citep{kullback1951information}.




\subsection{Merging Performance}
\label{sec:mperformance}
In this section, we present a comprehensive evaluation of our approach in comparison to state-of-the-art task vector merging methods, assessing its superiority across several fundamental aspects: Multi-task merging performance, generalization to unseen tasks and heterogeneous architectures.

\textbf{Substantially Higher Merging Performance.} The main results of merging performance of ViT-B/32 models on eight tasks are presented in this section, detailed \footnote{Please refer to the Appendix for experimental details, including the full list of tasks, datasets, and baselines.} in Table \ref{tab:mergemain}. We present two levels of granularity: Task-Wise (TW) and Layer-Wise (LW). Our method \method~ achieved an average accuracy (Avg Acc) of $76.4\%$ and $94.5\%$ in both TW and LW (\methodpp~) levels, outperforming the state-of-the-art (SOTA) method AdaMerging++ and WEMoE by a large margin of $2.7\%$ and $5.1\%$. While finer granularity is generally associated with improved merging performance~\citep{yang2023adamerging}, our \textbf{LW \methodpp~}, operating at the Layer-wise level, surpasses EMR-Merging~\citep{huang2024emr} which is based on the finer Parameter-wise granularity. We attribute the improvements to the ability of \learner~ to 
adapt weight coefficients based on their weight statistics to the merged model. In addition, the use of pseudo labels from task-specific teachers $\{T_{1}, T_{2}, \ldots, T_{k}\}$ provides stronger signals for \learner~ to better assign weight coefficients $\lambda$ than the entropy minimization approach in the AdaMerging++.

\begin{table}[h]
\centering
\footnotesize
\label{tab:multitask-merging}
\begin{tabular}{l|cccccccc|p{20pt}}
\toprule
\textbf{Method} & SU & CA & RE & EU & SV & GT & MN & DT & \textbf{Avg Acc} \\
\midrule
Pre-Trained & 62.3 & 59.7 & 60.7 & 45.5 & 31.4 & 32.6 & 48.5 & 43.8 & 48.0 \\
Individual & 75.3 & 77.7 & 96.1 & 99.7 & 97.5 & 98.7 & 99.7 & 79.4 & 90.5 \\
Traditional MTL & 73.9 & 74.4 & 93.9 & 98.2 & 95.8 & 98.9 & 99.5 & 77.9 & 88.9 \\
\midrule
\midrule
& \multicolumn{8}{c}{\textbf{Task-Wise}} \\
\midrule
Weight Averaging & \underline{65.3} & 63.4 & 71.4 & 71.7 & 64.2 & 52.8 & 87.5 & 50.1 & 65.8 \\
Task Arithmetic & 55.2 & 54.9 & 66.7 & 78.9 & 80.2 & 69.7 & \underline{97.3} & 50.4 & 69.1 \\
Fisher Merging & \textbf{68.6} & \underline{69.2} & 70.7 & 66.4 & 72.9 & 51.1 & 87.9 & \textbf{59.9} & 68.3 \\
RegMean & \underline{65.3} & 63.5 & \textbf{75.6} & 78.6 & 78.1 & 67.4 & 93.7 & 52.0 & 71.8 \\

Ties-Merging & 59.8 & 58.6 & 70.7 & 79.7 & 86.2 & 72.1 & \textbf{98.3} & \underline{54.2} & 72.4 \\
TW AdaMerging & 58.0 & 53.2 & 68.8 & \textbf{85.7} & 81.1 & \textbf{84.4} & 92.4 & 44.8 & 71.1 \\
TW AdaMerging++ & 60.8 & 56.9 & 73.1 & 83.4 & \underline{87.3} & 82.4 & 95.7 & 50.1 & \underline{73.7} \\
\rowcolor{gray!15} \textbf{TW \method~ (Ours)} & 61.3 & \textbf{70.0} & \underline{74.2} & \underline{85.2} & \textbf{87.5} & \underline{82.5} & {96.2} & \underline{54.2} & \textbf{76.4} \\
\midrule

& \multicolumn{8}{c}{\textbf{Layer-Wise}} \\
\midrule
LW AdaMerging & 64.5 & 68.1 & 79.2 & {93.8} & 87.0 & {91.9} & {97.5} & 59.1 & 80.1 \\
LW AdaMerging++ & 66.6 & 68.3 & {82.2} & {94.2} & {89.6} & 89.0 & {98.3} & {60.6} & {81.1} \\
LW AdaMerging w/ Surgery & 69.8 & 71.0 & 88.9 & {98.1} & 91.7 & 96.5 & \underline{98.8} & 73.6 & 86.1 \\

LW AdaMerging w/ SurgeryV2 & \underline{74.7} & 71.4 & \textbf{95.1} & \textbf{99.6} & \textbf{96.8} & \textbf{98.9} & \textbf{99.6} & \underline{78.3} & 89.3 \\

WEMoE & {74.1} & \underline{77.4} & \underline{93.7} & \underline{99.1} & \underline{96.2} & \textbf{98.9} & \textbf{99.6} & {76.4} & \underline{89.4} \\



\rowcolor{gray!15} \textbf{LW \method~ (Ours)} & 67.4 & {74.1} & {82.9} & 91.1 & {89.8} & {94.7} & {98.3} & {77.5} & {84.5} \\

\rowcolor{gray!15} \textbf{LW \methodpp~ (Ours)} & \textbf{92.4} & \textbf{95.4} & \textbf{95.1} & {92.9} & {94.6} & \underline{98.7} & {98.5} & \textbf{88.4} & \textbf{94.5} \\



\midrule

& \multicolumn{8}{c}{\textbf{Parameter-wise}} \\
\midrule
EMR-MERGING & {75.2} & 72.8 & {93.5} & {99.5} & {96.9} & {98.1} & {99.6} & 74.4 & {88.7} \\

\bottomrule

\end{tabular}
\label{tab:mergemain}
\caption{Multi-task merging performance (Avg Acc $\%$) when merging ViT-B/32 models on eight tasks. Results of our method \method~ are shaded in gray. Bold and underscore indicate the highest and second-highest scores in each column within each group under Task-wise and Layer-wise settings. TW: Task-wise. LW: Layer-wise. PW: Parameter-wise.} 
\end{table}








It is worth noting that \method~ outperforms the Individual setting. The observed improvements can be attributed to learning (1) from a broader diversity of scenes and (2) the implicit noise patterns across the aggregated data \citep{yang2024model}, which \learner~ effectively leverages to enhance cross-domain generalization and surpass the performance of the teacher models \citep{nagarajan2023student, starodubcev2024your}.


\noindent \textbf{Significantly Enhanced Generalization.} A merged model is expected to generalize to unseen tasks by strategically transferring the knowledge from the combined set of old tasks. We benchmark such generalization ability of \method~ against four strong baselines: Task Arithmetic, Ties-Merging, AdaMerging, and AdaMerging++. We follow the same evaluation protocol in AdaMerging training on two groups of tasks, each group consisting of six seen tasks, and testing on two unseen tasks.

\begin{table}[h]
\centering
\footnotesize
\label{tab:generalization}
\begin{tabular}{l|cccccc|p{40pt}||cc|p{40pt}}
\toprule
& \multicolumn{6}{c}{\textbf{Seen Tasks}} & & \multicolumn{3}{c}{\hfill\textbf{Unseen Tasks}} \\
\toprule
\textbf{Method} & SU & CA & RE & DT & SV & GT & \textbf{Avg Acc} & MN & EU & \textbf{Avg Acc} \\
\midrule
Task Arithmetic & 63.3 & 62.4 & 75.1 & 57.8 & 84.6 & 80.4 & 70.6 & 77.2 & 46.2 & 61.7 \\
Ties-Merging & 67.8 & 66.2 & 77.2 & 56.7 & 77.1 & 70.9 & 69.3 & 75.9 & 43.3 & 59.6 \\
AdaMerging & 65.2 & 65.9 & \textbf{88.5} & 61.1 & 92.2 & \underline{91.5} & 77.4 & \underline{84.0} & \underline{56.1} & \underline{70.0} \\
AdaMerging++ & \underline{68.2} & \underline{67.6} & 86.3 & \underline{63.6} & \underline{92.6} & 89.8 & \underline{78.0} & 83.9 & 53.5 & 68.7 \\
\rowcolor{gray!15} \textbf{\method~} & \textbf{69.1} & \textbf{71.3} & \underline{86.7} & \textbf{75.2} & \textbf{93.2} & \textbf{95.7} & \textbf{81.9 (+3.9)} & \textbf{85.1} & \textbf{56.4} & \textbf{70.8 (+0.8)} \\
\midrule
\textbf{Method} & SU & CA & GT & EU & DT & MN & \textbf{Avg Acc} & RE & SV & \textbf{Avg Acc} \\
\midrule
Task Arithmetic & 64.0 & 64.0 & 75.2 & 87.7 & 57.0 & 95.7 & 73.9 & 52.3 & 44.9 & 51.1 \\
Ties-Merging & 68.0 & 67.1 & 67.7 & 78.4 & 56.5 & 92.8 & 71.8 & \textbf{58.7} & 49.2 & 53.9 \\
AdaMerging & 67.1 & 67.8 & \underline{94.8} & \underline{94.4} & 59.6 & 98.2 & 80.3 & 50.2 & 60.9 & 55.5 \\
AdaMerging++ & \underline{68.9} & \underline{69.6} & 91.6 & 94.3 & \underline{61.9} & \underline{98.7} & \underline{80.8} & 52.0 & \underline{64.9} & \underline{58.5} \\
\rowcolor{gray!15} \textbf{\method~} & \textbf{69.6} & \textbf{73.3} & \textbf{96.1} & \textbf{95.4} &  \textbf{74.1} & \textbf{97.2} & \textbf{84.3 (+3.5)} & \underline{54.2} & \textbf{67.1} & \textbf{60.7 (+2.2)} \\
\bottomrule
\end{tabular}
\label{tab:genunseen}
\caption{Generalization results (Avg Acc $\%$) on two unseen tasks when merging Layer-Wise ViT-B/32 models on six tasks. \method~: shaded in gray. Bold: top score. Underscore: 2nd-highest score.} 
\end{table}

Details are presented in Table \ref{tab:genunseen}, where in both groups our proposed \method~ achieved $70.8\%$ and $60.7\%$, significantly outperforming AdaMerging by $0.8\%$ and $2.2\%$. Such improvements are attributed to both (1) the careful feature design of weight statistics that captures the dominant information regarding weight distributions from pre-trained models, which potentially helps reduce noise from each task dataset; and (2) the joint training from all old tasks on the task-specific teacher-distilled labels, enabling the implicit learning of task-agnostic and task-specific features that can benefit the generalization ability.

\noindent \textbf{Extension to Heterogeneous Architectures.} Our \method~ offers the first and unique advantage without the assumption of architectural identity in prior works \citep{wortsman2022model, ilharcoediting, yadav2023ties, matena2022merging, jindataless}.
To verify the performance of varying architectures, we conduct experiments on ResNet50 (RN) and ViT-B/32 (VT) to represent Convolutional Neural Network (CNN) and Vision Transformer (ViT) architectures.

In particular, we distill fine-tuned VT teachers into a RN \citep{khanuja2021mergedistill} student on three diverse tasks of CIFAR-10 (CI), EuroSAT (EU) and Stanford Cars (CA) with the distillation loss: 
\begin{equation}
\mathcal{L}\;=\;
\alpha\,\mathcal{L}_{\mathrm{CE}}(y,\hat y)
\;+\;(1-\alpha)\,T^2\,\mathcal{L}_{\mathrm{KL}}\!\bigl(\sigma(\tfrac{z}{T}),\;\sigma(\tfrac{z_t}{T})\bigr),
\end{equation}
where $\mathcal{L}_{\mathrm{KL}}$ denotes KL-Divergence, $z$ is logit, $T=4.0$ represents temperature, $\alpha=0.7$ is the weight balance of two sub-losses. CI is used due to the available pre-trained RN weights. Remarkably, the distilled RN matches its VT teacher’s accuracy, achieving $76.4\%$ (VT: $77.7\%$) for CA and $94.5\%$ for EU (VT: $99.7\%$) despite the architectural difference shown in Table \ref{tab:hete}. We then apply our \method~ to combine the CI–trained RN and its distilled variants. We merge multiple task models into a single RN using the merging coefficients inferred by \learner~, yielding a $7.6\%$ average improvement over the vanilla Task-Arithmetic of $73.7\%$ and achieving $81.3\%$ average accuracy.


\begin{table*}[h] 
\footnotesize 
  \begin{center} 
    { 
\begin{tabular}{c|ccc|c}
\toprule
\textbf{Method} & CI & CA & EU  & \textbf{Avg Acc} \\
\midrule 
Backbone & RN & VI & VI & -\\ 
Distilled &  - & RN & RN & - \\
\midrule
Individual & 97.8 & 77.7 & 99.7 & 91.7\\
Distilled & - & 76.4 & 94.5  & - \\
\midrule
\midrule
Weight Averaging & 77.1 & 56.4 & 64.9 & 59.4 \\
\midrule
Ties-Merging & 76.5 & 52.8 & 80.1 & 69.8 \\
Task Arithmetic & 81.4 & 61.6 & 78.2 & 73.7 \\
\midrule
\rowcolor{gray!15} \textbf{LW \method~} & 87.2 & 68.4 & 88.4 &  81.3\\
\bottomrule
\end{tabular}
}\end{center} 
\caption{Multi-task merging performance (Avg Acc \%) of models in heterogeneous architectures: ResNet50 (RN) \& ViT-B/32 (VT). \method~: shaded in gray.} 
\label{tab:hete}
\end{table*} 

\subsection{\method~ Analysis}
\noindent
\begin{minipage}[t]{0.48\textwidth}
\raggedright
\vspace{-35pt} 
\textbf{Label \& Loss Function Study.}

We conduct a loss function study on ViT-B/32 (4) models merged from four tasks, as shown in Table 5. Observe that \method~ trained on pseudo labels via Task-Specific Teacher Distillation (KD) achieves similar performance to \method~ trained on ground-truth labels (GT), with $88.5\%$ and $81.2\%$ average accuracy in TW and $90.4\%$ and $83.5\%$ in LW levels.
\end{minipage}%
\hfill
\vspace{4pt}
\begin{minipage}[t]{0.48\textwidth}
\centering
\scriptsize
\begin{tabular}{ll|cccc|c}
\toprule
{Loss} & Level & CA & EU & RE & GT & \textbf{Avg Acc} \\
\midrule
GT & TW & 73.2 & 94.2 & 91.1 & 95.6 & 88.5 \\
KD & TW & 64.2 & 88.6 & 85.2 & 86.7 & 81.2 \\
\midrule
GT & LW & 75.6 & 96.3 & 92.1 & 97.6 & 90.4 \\
KD & LW & 68.7 & 91.6 & 87.2 & 93.5 & 83.5 \\
\bottomrule
\label{tab:loss}
\end{tabular}
{Table 5: Multi-task performance (Avg Acc \%) of \method~ when merging ViT-B/32 (4) models across four tasks. \method: shaded in gray. GT: Ground Truth. KD: Knowledge Distillation. TW: Task-wise. LW: Layer-wise.}
\end{minipage}

\noindent
\begin{minipage}[t]{0.46\textwidth}
\raggedright
\vspace{-35pt}
\textbf{Statistical Feature Ablation Study.}

We conduct an ablation study on the statistical features. Results in Table 6 show that combining all statistical features improves merging performance, validating our design choice. Notably, the singular values $\sigma^{\prime}$ improve the multi-task performance in both same and different architecture settings by $3.0\%$ and $3.2\%$ increase of average accuracy, justifying our design choice of using SVD.
\end{minipage}%
\hfill
\begin{minipage}[t]{0.52\textwidth}
\centering
\scriptsize
\begin{tabular}{p{3pt}p{1pt}p{1pt}p{5pt}|p{31pt}||p{3pt}p{1pt}p{1pt}p{5pt}|p{31pt}}
\toprule
\multicolumn{5}{c||}{\textbf{Same Architecture}} & \multicolumn{5}{c}{\textbf{Different Architecture}} \\
\cmidrule{1-5} \cmidrule{6-10}
$\mu_{\theta_{k}}$ & $\sigma^2$ & $m$ & $\sigma^{\prime}$ & \textbf{Avg Acc} &
$\mu_{\theta_{k}}$ & $\sigma^2$ & $m$ & $\sigma^{\prime}$ & \textbf{Avg Acc} \\
\midrule
\cmarkk &  &  &  & 83.4 & \cmarkk &  &  &   & 76.2 \\
\cmarkk & \cmarkk &  &  & 84.1 (+0.7) & \cmarkk & \cmarkk &  &  & 77.5 (+1.3) \\
\cmarkk & \cmarkk & \cmarkk &  & 87.2 (+3.1) & \cmarkk & \cmarkk & \cmarkk &  & 78.1 (+0.6) \\
\rowcolor{gray!15} \cmarkk & \cmarkk & \cmarkk & \cmarkk & \textbf{90.2 (+3.0)} & \cmarkk & \cmarkk & \cmarkk & \cmarkk & \textbf{81.3 (+3.2)} \\
\bottomrule
\end{tabular}
\label{tab:ablstudy}
{Table 6: Multi-task performance (Avg Acc \%) of \method~ when ablating statistical features of ViT-B/32 (4) models on four tasks: CA, EU, RE \& GT. Bold: top score. \method~: shaded in gray.}
\end{minipage}



\begin{figure}[h]
  \centering
    {\includegraphics[width=1\textwidth]{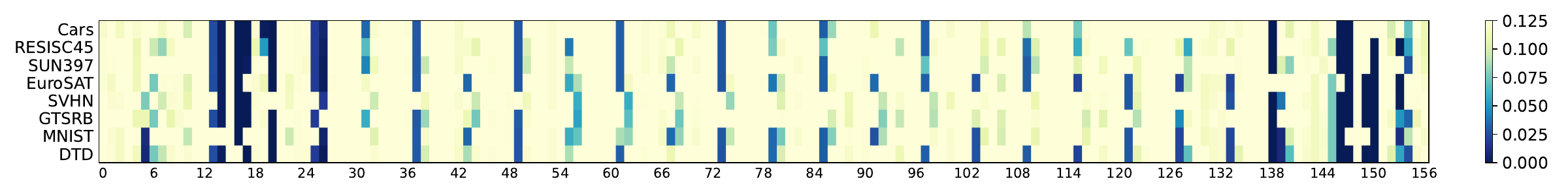}}
\vspace{-14pt}
  \caption{Heatmap of \method~ merging coefficients $\lambda$ of ViT-B/32 (4) across eight tasks. X-axis: layer index. Y-axis: Tasks. Coefficients are normalized to sum to 1.}
  \label{fig:lambda_coeff}
\end{figure}

\noindent \textbf{Coefficient Analysis.} We visualize the heatmap of ViT-B/32 (4) across eight tasks in Fig. \ref{fig:lambda_coeff}. We make several key observations: (1) the \textbf{common recurring pattern} of coefficients $\lambda$ across all eight tasks from earlier (left) to deeper (right) layers aligns with the repeated self-attention blocks in the ViT architecture, e.g. Multi-Head Self-Attention (MHSA), MLP (Feed-Forward Network), and LayerNorm, etc, demonstrating the need of various coefficients for various types of layers; (2) The \textbf{sparse non-uniform coefficient distributions} (various colors like Layer $13$, $19$ or $25$) suggests that merging layers can be more efficient at some specific layers instead of using one coefficient for an entire pre-trained model, justifying the our granularity choice of Layer-Wise over Task-Wise level; (3) some \textbf{task-specific coefficient distributions} verify the necessity of assigning distinct merging coefficients across tasks in various layers, such as in Layer 5 vs.\ 147. Such distributions reflect the various visual representations for different semantics learned across both layers and tasks. More visualizations are provided in the Appendix for in-depth analysis.


\vspace{-6pt}


\section{Conclusion}
Model merging offers a compelling post-hoc advantage to reduce memory storage from a corpus of large pre-trained models. We propose \method~, a novel merging technique guided by model weight statistical features learned through Task-Specific Teacher Distillation without relying on human annotated samples. The key intuition lies in the guidance of weight statistics using a lightweight MLP learner, dubbed \learner~, to infer merging coefficients. Comprehensive experiments demonstrate the effectiveness of our proposed \methodpp~ (extended version of \method~) in model mering from eight diverse tasks, achieving $94.5\%$ average accuracy and surpassing the state-of-the-art approach WEMoE ($89.4\%$) by a large margin of $5.1\%$.








\newpage

\newpage
\appendix

\section{Experiment Settings}
This section presents a comprehensive overview of the datasets, baseline methods, and training procedures.

\noindent \textbf{Task.} A task is referred to the specific problem or objective that a model is designed to solve. In this paper, a task is defined as classifying images within a given dataset.

\noindent \textbf{Dataset Details.} This study follows the multi-task model merging protocol from Task Arithmetic \citep{ilharcoediting}, Ties-Merging \citep{yadav2023ties} and AdaMerging \citep{yang2023adamerging} on eight image classification datasets. The details are provided below:

\begin{itemize}[nosep, leftmargin=1.5em]
    \item \textbf{SUN397 (SU)} \citep{xiao2016sun}: a scene classification dataset consisting of 397 classes and a total of 108,754 images, with each class containing a minimum of 100 images.

    \item \textbf{Stanford Cars (CA)} \citep{krause20133d}: a car classification benchmark dataset comprosing 196 categories and 16,185 images in total. For each category, the dataset is evenly divided into training and test sets in a 1:1 ratio.
    
    \item \textbf{RESISC45 (RE)} \citep{cheng2017remote}: a remote sensing image scene classification benchmark with 45 scene classes and 31,500 images. Approximately 700 images are included in each class.
    
    \item \textbf{EuroSAT (EU)} \citep{helber2019eurosat}: a 10-class satellite image classification dataset with 27,000 labeled and geo-referenced images.
    
    \item \textbf{SVHN (SV)} \citep{netzer2011reading}: a real-world digit classification dataset derived from house numbers in Google Street View images. This datasets consists of 10 classes with 73,257 training samples and 26,032 test samples. Additional 531,131 samples are available for training.
    
    \item \textbf{GTSRB (GT)} \citep{stallkamp2011german}: a traffic sign classification dataset consisting of 43 classes and more than 50,000 samples in total.
    
    \item \textbf{MNIST (MN)} \citep{lecun1998mnist}: a benchmark dataset for image classification, containing grayscale images of handwritten digits across 10 classes. It includes 60,000 training and 10,000 test images, with a balanced number across classes.
    
    \item \textbf{DTD (DT)} \citep{cimpoi2014describing}: a texture classification dataset consisting of 47 classes and a total of 5,640 images, with approximately 120 images per class.
\end{itemize}

\noindent \textbf{Baseline Details.}
We evaluate performance using eight comparison baselines and four alternative configurations of our method.

\begin{itemize}[nosep, leftmargin=*]

    \item \textbf{Individual}: Each task is handled by an independently fine-tuned model with no interference between tasks. However, this approach cannot perform multiple tasks simultaneously.

    \item \textbf{Traditional MTL}: This approach aggregates the original training data from all tasks to train a single multi-task model. It serves as a reference \textit{upper bound} for evaluating model merging performance.

    \item \textbf{Weight Averaging}: A simple model merging technique that averages the parameters of multiple models directly. It is typically considered a \textit{lower bound} for model merging performance.

    \item \textbf{Fisher Merging} \citep{matena2022merging}: This method computes the Fisher Information Matrix to assess parameter importance, guiding the model merging process based on these importance scores.

    \item \textbf{RegMean} \citep{jindataless}: This approach introduces a regularization constraint during merging, enforcing the $L_{2}$ distance between the merged model and individual models to remain small.

    \item \textbf{Task Arithmetic} \citep{ilharcoediting}: This method is the first to propose the concept of ``task vectors'' and merges these vectors into a pre-trained for model merging.

    \item \textbf{Ties-Merging} \citep{yadav2023ties}: This approach addresses task conflicts in Task Arithmetic \citep{ilharcoediting} by removing redundant parameters and resolving sign conflicts through a three-step procedure: Trim, Elect Sign, and Disjoint Merge.

    \item \textbf{EMR-MERGING} \citep{huang2024emr}: A tuning-free method that merges models in three steps, by selecting a unified parameter sign (Elect), aligning task-specific parameters via masking (Mask), and adjusting their magnitudes with task-specific scaling factors (Rescale).

    \item \textbf{AdaMerging} \citep{yang2023adamerging}: This method builds on Task Arithmetic \citep{ilharcoediting} by employing an unsupervised method to automatically learn merging coefficients for each task vector.

    \item \textbf{AdaMerging++} \citep{yang2023adamerging}: An extension of Ties-Merging \citep{yadav2023ties} that uses an unsupervised approach to learn task-specific merging coefficients.

    \item \textbf{AdaMerging w/ Surgery} \citep{yang2024representation}: A task-specific lightweight module that reduces representation bias through unsupervised optimization of merged-to-individual model epresentation alignment based on AdaMerging only at the final layer of the model.

    \item \textbf{AdaMerging w/ SurgeryV2}
    \citep{yang2024surgeryv2}: This framework extends Representation Surgery \citep{yang2024representation} to all intermediate layers through layer-wise transformations.

    \item \textbf{WEMoE} \citep{tang2024merging}: This approach merges most parameters and upscales Transformer MLP layers to a weight-ensembling mixture of experts (MoE) module.

    \item \textbf{\method~ (Ours)}: A lightweight learning-based method guided by the weight distribution statistical features (stats) of task-specific pre-trained weight models, including the mean, variance, magnitude and singular values. This method employs \learner~ to learn stats by knowledge distillation from task-specific teachers without manual labels from a small portion of validation data.

    \item \textbf{\methodpp~ (Ours)}: An extended version of \method~ trained on more validation data.

\end{itemize}

\noindent \textbf{Training Details.} We follow the same training procedure outlined in AdaMerging \citep{yang2023adamerging}.

\begin{itemize}[nosep, leftmargin=*]

    \item \textbf{Task-Specific Teacher}: For each task, we utilize its corresponding \textbf{Individual} model as the \textbf{Teacher}.

\end{itemize}

Code is available at \href{https://github.com/statsmerging/statsmerging}{https://github.com/statsmerging/statsmerging}.

\section{Extended Experiments}

\subsection{Robustness Evaluation}
We evaluate the robustness of \method~ against Task Arithmetic \citep{ilharcoediting} and AdaMerging~\citep{yang2023adamerging} under three image corruption scenarios: Motion Blur, Impulse Noise, and Gaussian Noise. The corrupted test sets are constructed following the protocols outlined in~\citep{yang2023adamerging, hendrycks2019robustness}. We assess performance on four datasets: Stanford Cars (CA)~\citep{krause20133d}, EuroSAT (EU)~\citep{helber2019eurosat}, RESISC45 (RE)~\citep{cheng2017remote}, and GTSRB (GT)~\citep{stallkamp2011german}. Results are reported in Table~\ref{tab:robust}. Overall, \method~ consistently outperforms the baselines. On the clean test set, it achieves a $2.4\%$ accuracy improvement over AdaMerging. Under corrupted conditions, \method~ yields performance gains of $3.1\%$, $6.3\%$, and $4.3\%$ for Motion Blur, Impulse Noise, and Gaussian Noise, respectively.

\begin{table}[h]
\centering
\caption{Robustness results when merging ViT-B/32 models on four tasks. \method~: shaded in gray. Bold: top score. Values are reported in \%.}
\label{tab:robustness}
\hspace*{-20pt}
\begin{tabular}{l|cccc|p{45pt}}
\toprule
\textbf{Method} & CA & EU & RE & GT & \textbf{Avg Acc} \\
\midrule
& \multicolumn{5}{c}{\textbf{Clean Test Set}} \\
\midrule
Task Arithmetic & 66.9 & 94.7 & 82.6 & 75.1 & 79.8 \\
AdaMerging & 73.7 & 96.1 & 85.8 & 96.3 & 88.0 \\
\rowcolor{gray!15} \textbf{\method~} & \textbf{75.6} & \textbf{96.3} & \textbf{92.1} & \textbf{97.6} & \textbf{90.4 (+2.4)} \\

\midrule

& \multicolumn{5}{c}{\textbf{Motion Blur}} \\
\midrule
Task Arithmetic & 65.3 & 68.1 & 80.0 & 64.2 & 69.4 \\
AdaMerging & 71.2 & 74.6 & 82.7 & 94.1 & 80.6 \\
\rowcolor{gray!15} \textbf{\method~} & \textbf{73.5} & \textbf{76.9} & \textbf{89.2} & \textbf{95.2} & \textbf{83.7 (+3.1)} \\

\midrule

& \multicolumn{5}{c}{\textbf{Impulse Noise}} \\
\midrule
Task Arithmetic & 62.1 & 49.1 & 72.7 & 40.4 & 56.1 \\
AdaMerging & 67.2 & 30.8 & 75.9 & 77.5 & 62.8 \\
\rowcolor{gray!15} \textbf{\method~} & \textbf{70.4} & \textbf{50.4} & \textbf{77.6} & \textbf{78.1}  & \textbf{69.1 (+6.3)} \\

\midrule

& \multicolumn{5}{c}{\textbf{Gaussian Noise}} \\
\midrule
Task Arithmetic & 63.6 & 55.4 & 75.9 & 49.4 & 61.1 \\
AdaMerging & 69.9 & 41.2 & 80.6 & 76.0 & 66.9 \\
\rowcolor{gray!15} \textbf{\method~} & \textbf{71.2} & \textbf{53.6} & \textbf{82.1} & \textbf{78.0} & \textbf{71.2 (+4.3)} \\

\bottomrule
\end{tabular}
\label{tab:robust}
\end{table}

\newpage
\subsection{Label Type and Loss Function Analysis}
In this section, we analyze the performance of training \learner~ on two types of pseudo labels: (1) Soft Pseudo Labels, and (2) Hard Pseudo Labels, the former of which is commonly employed in knowledge distillation frameworks \citep{gou2021knowledge, hinton2015distilling} especially for classification tasks. Formally, we present two versions of our training losses:

\noindent \textbf{Soft Pseudo Labels (SPL):} The predicted class probability distribution. Thus we use Kullback–Leibler divergence (KL-Div) \citep{kullback1951information} loss function:
\begin{equation}
    \mathcal{L}_{\mathrm{KL}} = \sum_{c=1}^{C_{m}} p_{c,k} \log \left( \frac{p_{c,k}}{q_{c}} \right)
\end{equation}
where $p_{c,k}$ is the predicted probability of class $c$ from the pre-trained model $\theta_{k}$ on task $k$, and $q_{c}$ is the predicted probability of class $c$ from the merged model $\theta_{m}$.

\noindent \textbf{Hard Pseudo Labels (HPL):} The predicted class label in one-hot encoded format. Therefore, the cross-entropy loss is applied:
\begin{equation}
    \mathcal{L}_{\mathrm{CE}} =  - \sum^{C_{m}}_{c=1} \hat{y}_{c,k} \log(\hat{y}_{c}))
\end{equation}

Results are shown in \ref{tab:pseudotypes}. 
We highlight two key observations: (1) Training \learner~ with Hard Pseudo Labels (HPL) using cross-entropy loss (KD CE) yields performance comparable to training with ground-truth labels (GT CE), achieving $81.2\%$ vs. $88.5\%$ at the task-wise (TW) level and $83.5\%$ vs. $90.4\%$ at the layer-wise (LW) level. Importantly, \method~ eliminates the need for manually annotated labels, validating our intuition of leveraging task-specific teacher knowledge for supervision. (2) When trained on Soft Pseudo Labels (SPL) using KL-Divergence loss (KL-Div), \learner~ underperforms relative to HPL with cross-entropy, obtaining $73.3\%$ vs. $81.2\%$ at the TW level and $52.4\%$ vs. $83.5\%$ at the LW level, respectively. 


We hypothesize that the observed performance drop is due to noisy inter-class relationships within the aggregated dataset ~\citep{yuan2021rethink}. While a detailed investigation of these relationships is beyong the scope of this work on model merging, we believe it presents promising directions for future research.

Identifies "regularization samples" where soft labels degrade performance due to poor calibration or noisy class relationships. Proposes weighted soft labels to mitigate these issues.

\begin{table}[h]
\centering
\caption{Multi-task performance (Avg Acc $\%$) of \method~ when merging ViT-B/32 (4) models on four tasks. \method~: shaded in gray. GT: Ground Truth. KD: Knowledge Distillation. GL: Granularity level. TW: Task-wise. LW: Layer-wise.}
\label{tab:multitask-merging}
\hspace*{-30pt}
\begin{tabular}{ll|cccc|c}
\toprule
GL & {Loss} & CA & EU & RE & GT & \textbf{Avg Acc} \\
\midrule
TW & GT CE & 73.2 & 94.2 & 91.1 & 95.6 & 88.5 \\
TW  & KD KL-Div & 56.5 & 97.6 & 56.5 & 82.4 & 73.3 \\
\rowcolor{gray!15} TW  & KD CE & 64.2 & 88.6 & 85.2 & 86.7 & 81.2 \\
\midrule
LW  & GT CE & 75.6 & 96.3 & 92.1 & 97.6 & 90.4 \\
LW  & KD KL-Div & 53.1 & 41.4 & 65.9 & 49.1 & 52.4 \\
\rowcolor{gray!15} LW  & KD CE & 68.7 & 91.6 & 87.2 & 93.5 & 83.5 \\

\bottomrule
\end{tabular}
\label{tab:pseudotypes}
\end{table}

\newpage
\subsection{Efficient Inference}

\learner~ is designed to be lightweight, introducing minimal spatial and computational overhead to the overall merging process. As shown in Table~\ref{tab:eff}, it contains only 10.99M parameters, requires 2.95 GFLOPs, and achieves an inference time of 5.26\,ms on an NVIDIA RTX A6000 GPU.

\begin{table}[h]
\centering
\caption{Model Size and Computational Overhead of \learner~.}
\label{tab:efficiency}
\hspace*{-30pt}
\begin{tabular}{ccc}
\toprule
$\#$Params (M) & GFLOPs & Inference Time (ms) \\
\midrule
 10.99 & 2.95 &  5.26 \\
\bottomrule
\end{tabular}
\label{tab:eff}
\end{table}

\subsection{Training Curve}
\begin{figure}[h]
\hspace*{-42pt}
  \centering
        {
    \includegraphics[width=\textwidth]{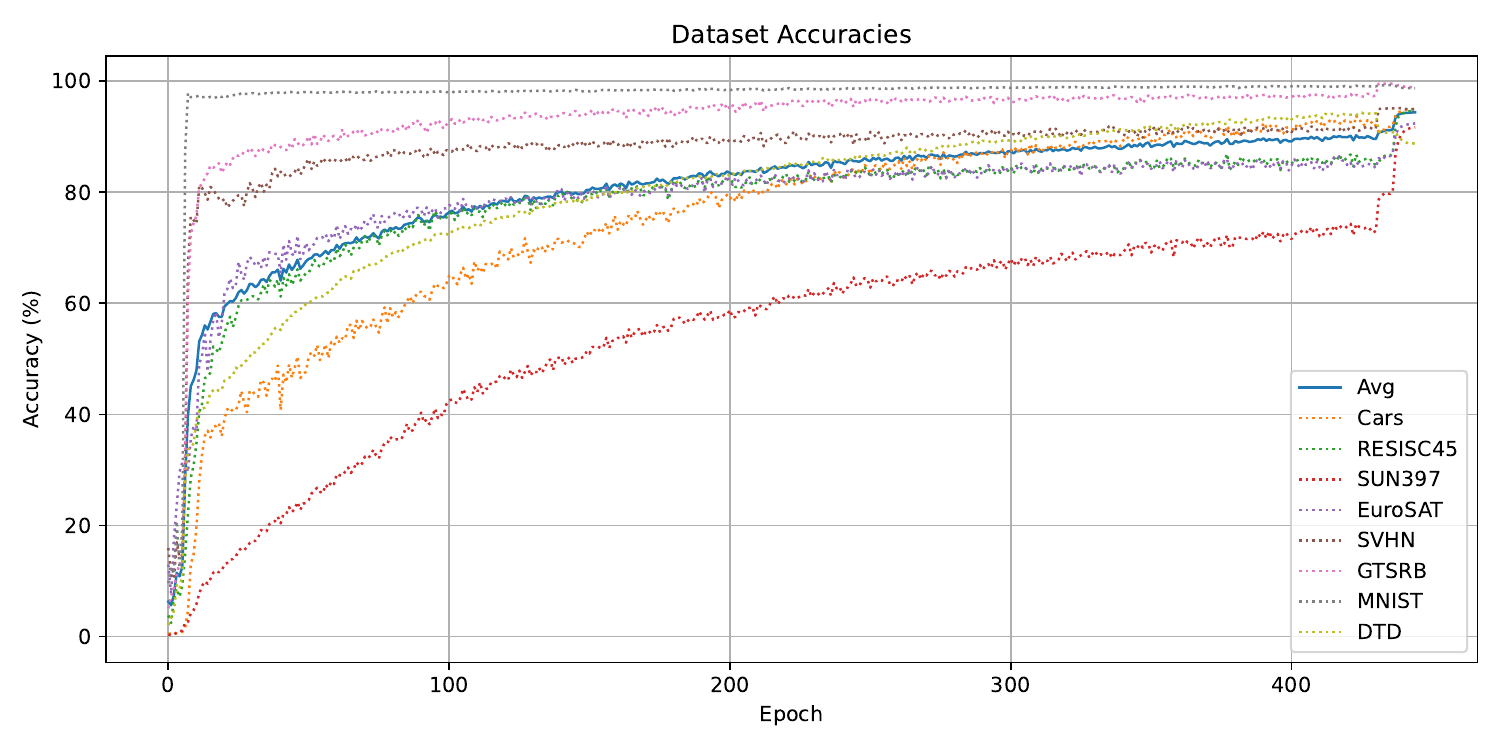}}
  \caption{\methodpp~ Training Accuracy Curve.}
  \label{fig:traincurve}
\end{figure}

The training curve is shown in Figure \ref{fig:traincurve}.

\subsection{Future Work and Limitations}

In this work, we focus on vision-based classification tasks, leaving extensions to other domains, such as object detection \citep{tan2020efficientdet}, super-resolution \citep{sun2022shufflemixer}, and image and video restoration \citep{liang2021swinir, merugu2025joint}, for future work. Additionally, expanding this approach to language tasks, particularly large language models (LLMs) ~\citep{yang2024model, song2024hierarchical, zhang2024cam, tie2025survey, kallini2025mrt5}, as well as to multi-modal learning \citep{zhu2025remedy, du2025adamms, bousselham2024grounding, lin2024sphinx}, represents a promising direction for further research.

\bibliographystyle{plainnat}
\section*{}
{
\small
\bibliography{main}
}
\end{document}